\documentclass[10pt,twocolumn,letterpaper]{article}

\usepackage{cvpr}
\usepackage{times}
\usepackage{epsfig}
\usepackage{graphicx}
\usepackage{amsmath}
\usepackage{array}
\usepackage{amssymb}
\usepackage{amsbsy}
\usepackage{booktabs}
\usepackage{enumitem}
\usepackage{mathtools}
\usepackage{subfigure}
\usepackage{tabularx}
\usepackage{multirow}
\usepackage{caption}
\usepackage{overpic}
\usepackage{xcolor}
\usepackage{authblk}
\usepackage[title]{appendix}
\usepackage[hang,flushmargin]{footmisc}

\DeclarePairedDelimiterX{\infdivx}[2]{(}{)}{%
  #1\;\delimsize\|\;#2%
}
\DeclarePairedDelimiterX{\inp}[2]{\langle}{\rangle}{#1, #2}
\newcommand{\tpm}{$\ \pm\ $}
\newcommand{\KLD}{\textrm{KL}\infdivx}

\DeclareMathOperator*{\argmin}{\arg\!\min}

\newcommand{\conv}[1]{\textrm{CONV}_{#1}}
\newcommand{\convblock}[1]{\textrm{CONVBlock}_{#1}}

\newcommand{\relu}{\textrm{LReLU}}
\newcommand{\ds}{\textrm{DS}_{2}}
\newcommand{\us}{\textrm{US}_{2}}
\newcommand{\res}[1]{\textrm{ResBlock}_{#1}}
\newcommand{\cres}[1]{\textrm{CResBlock}_{#1}}

\newcommand{\fc}[1]{\textrm{FC}_{#1}}

\newcommand{\myparagraph}[1]{\vspace{0.38em}\noindent\textbf{#1}}
\newlength{\oldparindent}
\newcommand{\app}{CAPE\xspace}
\renewcommand{\etal}{et al.}
\renewcommand{\ie}{i.e.~}
\renewcommand{\eg}{e.g.~}

\newcommand{\websiteCAPE}{\mbox{\url{https://cape.is.tue.mpg.de}}}

\makeatletter
\renewcommand\AB@affilsepx{\quad\protect\Affilfont}
\makeatother

\usepackage[pagebackref=true,breaklinks=true,letterpaper=true,colorlinks,bookmarks=false]{hyperref}

\cvprfinalcopy 

\def\cvprPaperID{00382} 

\newcommand\blfootnote[1]{%
  \begingroup
  \renewcommand\thefootnote{}\footnote{#1}%
  \addtocounter{footnote}{-1}%
  \endgroup
}
\ifcvprfinal\fi
\begin{document}

\title{Learning to Dress 3D People in Generative Clothing}
\author[1]{Qianli Ma}
\author[1]{Jinlong Yang}
\author[1,2]{Anurag Ranjan}
\author[4]{Sergi Pujades}
\author[5]{Gerard Pons-Moll}
\author[3]{\\Siyu Tang\textsuperscript{*}}
\author[1]{Michael J. Black}
\affil[1]{\normalsize Max Planck Institute for Intelligent Systems, T\"ubingen, Germany} 
\affil[2]{\normalsize University of T\"ubingen, Germany \authorcr}
\affil[3]{\normalsize ETH Z\"urich, Switzerland}
\affil[4]{\normalsize Universit\'e Grenoble Alpes, Inria, CNRS, Grenoble INP, LJK, France \authorcr}
\affil[5]{\normalsize Max Planck Institute for Informatics, Saarland Informatics Campus, Germany \authorcr
  {\tt \small \{qma,~jyang,~aranjan,~black\}@tue.mpg.de \authorcr 
  {\tt \small sergi.pujades-rocamora@inria.fr \quad gpons@mpi-inf.mpg.de \quad siyu.tang@inf.ethz.ch}}
  }

\twocolumn[{%
\renewcommand\twocolumn[1][]{#1}%
\maketitle
\begin{center}
    \newcommand{\teaserwidth}{\textwidth}
\vspace{-0.4in}
    \centerline{
 \begin{overpic}[width=0.95\textwidth]{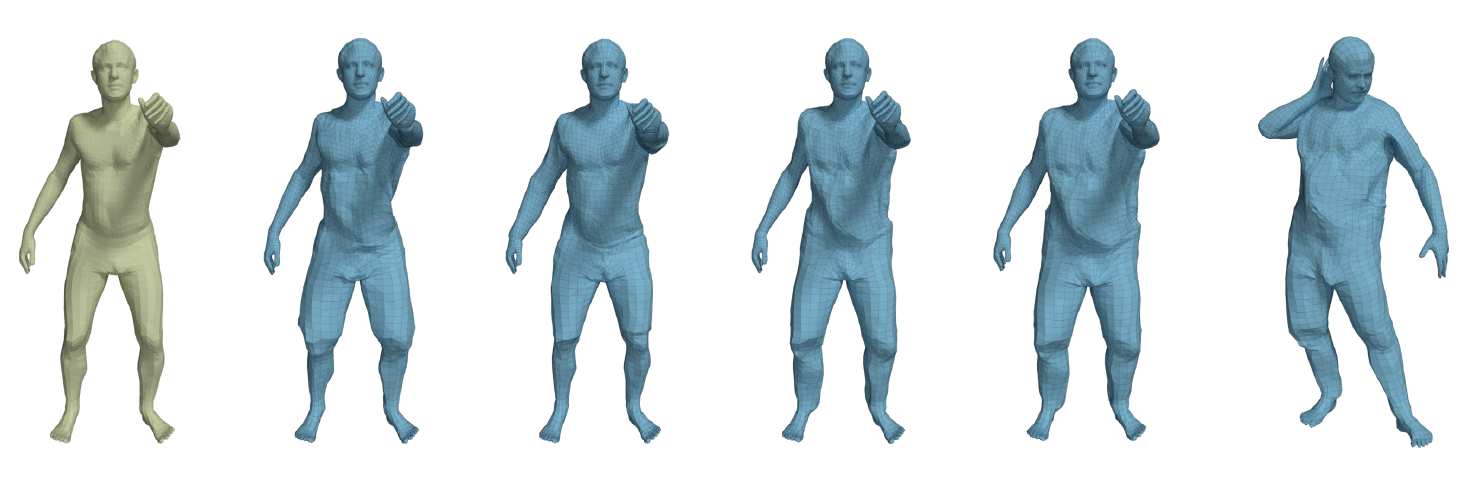}
\put(7,1.75){\textbf{(a)}}
\put(23.5,1.75){\textbf{(b)}}
\put(40.5,1.75){\textbf{(c)}}
\put(57,1.75){\textbf{(d)}}
\put(73.5,1.75){\textbf{(e)}}
\put(91,1.75){\textbf{(f)}}
\end{overpic}
     }
   \vspace{-0.1in}
   \captionof{figure}{\textbf{\app} model for clothed humans. Given a
     SMPL body shape and pose \textbf{(a)}, CAPE adds clothing by randomly sampling from a learned model \textbf{(b,~c)}, can generate different clothing types --- shorts in \textbf{(b,~c)} vs. long-pants in {\bf ~(d)}. The generated clothed humans can generalize to diverse body shapes {\bf (e)} and body poses {\bf (f)}. Best viewed zoomed-in on screen.}
\label{fig:teaser}
\end{center}%
}]
\maketitle
\thispagestyle{empty}

\begin{abstract}
\vspace{-0.06in}
Three-dimensional human body models are widely used in the analysis of
human pose and motion. Existing models, however, are learned from minimally-clothed 3D scans and thus do not generalize to the complexity of dressed people in common images and videos. Additionally, current models lack the expressive power needed to represent the complex non-linear geometry of pose-dependent clothing shapes. To address this, we learn a generative 3D mesh model of clothed people from 3D scans with varying pose and clothing. Specifically, we train a conditional Mesh-VAE-GAN to learn the clothing deformation from the SMPL body model, making clothing an additional term in SMPL. Our model is conditioned on both pose and clothing type, giving the ability to draw samples of clothing to dress different body shapes in a variety of styles and poses. To preserve wrinkle detail, our Mesh-VAE-GAN extends patchwise discriminators to 3D meshes. Our model, named CAPE, represents global shape and fine local structure, effectively extending the SMPL body model to clothing. To our knowledge, this is the first generative model that directly dresses 3D human body meshes and generalizes to different poses. The model, code and data are available for research purposes at \websiteCAPE.
\blfootnote{$^*$Work was done when S. Tang was at MPI-IS and University of T\"ubingen.}
\end{abstract}

\vspace{-0.8em}
\section{Introduction}
Existing generative human
models~\cite{anguelov2005scape, joo2018total, loper2015smpl,pavlakos2019expressive}
successfully capture the statistics of human shape and pose deformation, but still miss an important component: clothing.
This leads to several problems in various applications.
For example, when body models are used to generate synthetic training data~\cite{Hoffmann:GCPR:2019, multihumanflow, ranjan2018learning, varol2017learning}, the minimal body geometry results in a significant domain gap between synthetic and real images of humans.
Deep learning methods reconstruct human shape from images, based on minimally dressed human models~\cite{alp2018densepose,kanazawa2018end,SPIN:ICCV:2019,kolotouros2019convolutional,lassner2017unite,omran2018neural,pavlakos2019expressive,pavlakos2018learning}. Although the body pose matches the image observation, the minimal body geometry does not match clothed humans in most cases. These problems motivate the need for a parametric clothed human model.

\begin{figure*}[ht]
\centering
\begin{overpic}[width=0.9\textwidth]{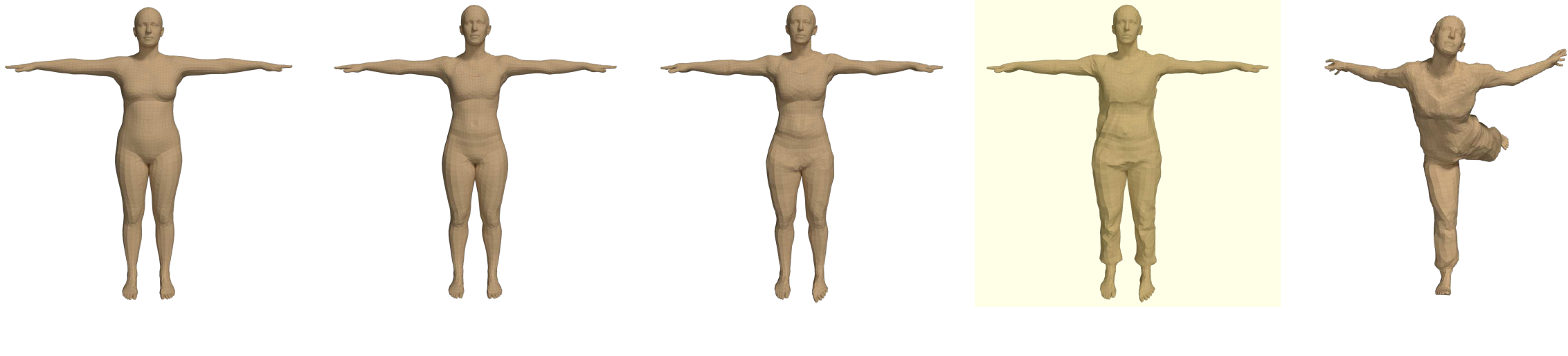}
\put(7.5,0.3){\scriptsize $\textbf{(a)}~\bar{T}$}
\put(22.5,0.3){\scriptsize $\textbf{(b)}~T(\beta)\!=\!\bar{T}\! + \!\mathit{B_S}(\beta)$}
\put(41,0.3){\scriptsize $\textbf{(c)}~T(\beta,\!\theta)\!=\! T(\beta)\!+\!\mathit{B_P}(\theta)$}
\put(62.3,0.3){\scriptsize $\textbf{(d)}~T_\textrm{clo}\!=\!T(\beta,\!\theta)\!+\!S_\textrm{clo}(z,\! \theta,\! c)$}
\put(85,0.3){\scriptsize $\textbf{(e)}~W(T_\textrm{clo},\! J(\beta),\! \theta,\! \mathcal{W})$}
\end{overpic}
\caption{Additive clothed human model. Our contribution is highlighted with yellow background. Following SMPL, our model \textbf{(a)} starts from a template mesh, and linearly adds offsets contributed by \textbf{(b)} individual body shape $\beta$, and \textbf{(c)} pose $\theta$; note the deformation on the hips and feet caused by the ballet pose. \textbf{(d)} We further add a clothing layer parametrized by pose $\theta$, clothing type $c$ and a clothing shape variable $z$. \textbf{(e)} The vertices are then posed using the skinning equation of SMPL.}\label{fig:smpl}
\vspace{-0.6em}
\end{figure*}
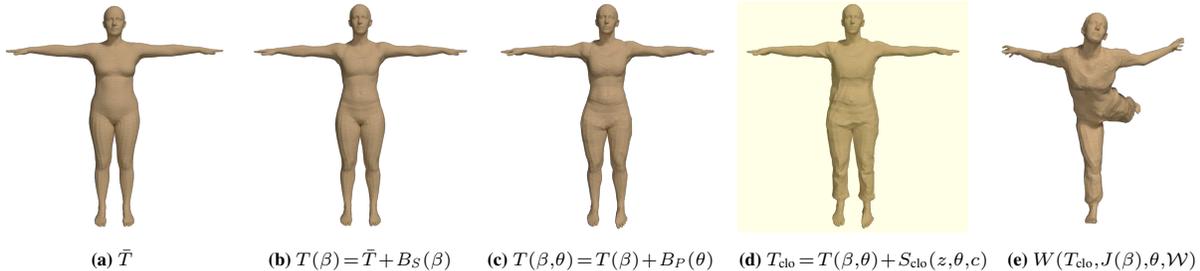

Our goal is to create a generative model of clothed human bodies that is low-dimensional, easy to pose, differentiable, can represent different clothing types on different body shapes and poses, and produces geometrically plausible results.
To achieve this, we extend SMPL \cite{loper2015smpl} and factorize clothing shape from the undressed body, treating clothing as an additive displacement in the canonical pose (see Fig.~\ref{fig:smpl}).
The learned clothing layer is compatible with the SMPL body model by design, enabling easy re-posing and animation.
The mapping from a given body shape and pose to clothing shape is
one-to-many. However, existing regression-based clothing
models~\cite{guan2012drape, yang2018analyzing} produce
\textit{deterministic} results that fail to capture the stochastic
nature of clothing deformations.
In contrast, we formulate clothing modeling
as a \textit{probabilistic generation task}: for a single pose and body shape, multiple clothing deformations can be sampled. 
Our model, called CAPE for ``Clothed Auto Person Encoding'', is conditioned on clothing types and body poses, so that it captures different types of clothing, and can generate \textit{pose-dependent deformations}, which are important for realistically modeling clothing.

We illustrate the key elements of our model in Fig.~\ref{fig:teaser}.
Given a SMPL body shape, pose and clothing type, \app can generate different structures of clothing by sampling a learned latent space. 
The resulting clothing layer plausibly adapts to different body shapes and poses.

\myparagraph{Technical approach.}
We represent clothing as a displacement layer using a graph that inherits the topology of SMPL. Each node in this graph represents the 3-dimensional offset vector from its corresponding vertex on the underlying body.
To learn a generative model for such graphs, we build a graph convolutional neural network (Sec.~\ref{sec:cape}), under the framework of a VAE-GAN~\cite{bao2017cvae,larsen2016autoencoding}, using graph convolutions~\cite{defferrard2016convolutional} and mesh sampling~\cite{ranjan2018generatingcoma} as the backbone layers.
This addresses the problem with existing generative models designed
for 3D meshes of human
bodies~\cite{litany2017deformable,verma2018feastnet} or
faces~\cite{ranjan2018generatingcoma} that tend to produce
over-smoothed results;  such smoothing is problematic for clothing where local details such as wrinkles matter.
Specifically, the GAN~\cite{goodfellow2014gan} module in our system encourages visually plausible wrinkles.
We model the GAN using a patch-wise discriminator for mesh-like graphs, and show that it effectively improves the quality of the generated fine structures.

\myparagraph{Dataset.} We introduce a dataset of 4D captured people performing a variety of pose sequences, in different types of clothing (Sec.~\ref{sec:dataset}). Our dataset consists of over 80K frames of 8 male and 3 female subjects captured using a 4D scanner. We use this dataset to train our network, resulting in a parametric generative model of the clothing layer.

\myparagraph{Versatility.} \app is designed to be ``plug-and-play'' for many applications that already use SMPL.
Dressing SMPL with \app yields 3D meshes of people in clothing, which can be used  for several applications such as generating training data, parametrizing body pose in a deep network, having a clothing ``prior'', or as part of a generative analysis-by-synthesis approach \cite{Hoffmann:GCPR:2019,ranjan2018learning,varol2017learning}.
We demonstrate this on the task of image fitting by extending SMPLify~\cite{bogo2016smplify} with our model.
We show that using CAPE together with SMPLify can improve the quality of reconstructed human bodies in clothing.

In summary, our key contributions are:
(1) We propose a probabilistic formulation of clothing modeling. 
(2) Under this formulation, we learn a conditional Mesh-VAE-GAN that captures both global shape and local detail of a mesh, with controlled conditioning based on human pose and clothing types.
(3) The learned model can generate pose-dependent deformations of clothing, and generalizes to a variety of garments.
(4) We augment the SMPL 3D human body model with our clothing model, and show an application of the enhanced ``clothed-SMPL''.
(5) We contribute a dataset of 4D scans of clothed humans performing a variety of motion sequences. 
Our dataset, code, and trained model are available for research purposes at \websiteCAPE.

\section{Related Work}
\begin{table*}[ht]
\small
\centering
\caption{Selection of related methods. Two main 3D clothing method classes exist: (1) Image-based reconstruction and capture methods, and (2) Clothing models that predict deformation as a function of pose.
Within each class, methods differ according to the criteria in the columns.}
\label{table:related_works}
\begin{tabular}{p{17.8mm}p{26.7mm}ccccccc}
\toprule
\multirow{2}{*}{\bf Method Class} & \multirow{2}{*}{\bf Methods}  & {\bf Parametric}  & {\bf Pose-dep.} &  {\bf Full-body} & {\bf Clothing} &  {\bf Captured} & {\bf Code} & {\bf Probabilistic}\\
 &  & {\bf Model}  & {\bf Clothing} & {\bf Clothing} & {\bf Wrinkles} & {\bf Data$^{*}$} & {\bf Public} & {\bf Sampling}\\
\hline
Image& Group 1 $^\dagger$  &No & No & Yes & Yes & Yes & Yes & No\\
Reconstruction & Group 2 $^\ddagger$  &Yes & No & Yes & Yes  & Yes & Yes & No\\
\hline
Capture &ClothCap~\cite{pons2017clothcap} & Yes & No & Yes & Yes & Yes & No & No\\
\hline
  &DeepWrinkles~\cite{lahner2018deepwrinkles} & Yes & Yes & No & Yes & Yes  & No & No\\
 &Yang \etal~\cite{yang2018analyzing} & Yes &  Yes & Yes & No & Yes & No & No\\
Clothing &Wang \etal~\cite{wang2018learning} & Yes &  No & No & No & No & Yes & Yes\\
Models &DRAPE~\cite{guan2012drape} & Yes &  Yes & Yes & Yes & No & No & No\\
&Sanesteban \etal~\cite{santesteban2019learning} & Yes &  Yes & Yes & Yes & No & No & No\\
&GarNet~\cite{gundogdu2019garnet} & Yes &  Yes & Yes & Yes & No & No & No\\
\hline
&\textbf{Ours} & Yes &  Yes & Yes & Yes & Yes & Yes & Yes\\
\bottomrule
\end{tabular}
\captionsetup{justification=raggedright,singlelinecheck=false}
\vspace{5pt}
\caption*{\footnotesize {* As opposed to simulated / synthetic data. \\
$^\dagger$ {Group 1}: BodyNet~\cite{varol2018bodynet}, DeepHuman~\cite{zheng2019deephuman}, SiCloPe~\cite{Natsume_2019_CVPR}, PIFu~\cite{saito2019pifu}, MouldingHumans~\cite{gabeur2910moulding}}. \qquad
$\ddagger$ {Group 2}: Octopus~\cite{alldieck19cvpr}, MGN~\cite{bhatnagar2019mgn}, Tex2Shape~\cite{alldieck2019tex2shape}.\\
}
\vspace{-1.5em}
\end{table*}

The capture, reconstruction and modeling of clothing has been widely studied.
Table~\ref{table:related_works} shows recent methods categorized into two major classes: (1) reconstruction and capture methods, and (2) parametric models, detailed as follows.

\myparagraph{Reconstructing 3D humans.}
Reconstruction of 3D humans from 2D images and videos is a classical computer vision problem.
Most approaches \cite{bogo2016smplify,guan2009estimating,kanazawa2018end,SPIN:ICCV:2019,kolotouros2019convolutional,lassner2017unite,omran2018neural, pavlakos2019expressive, smith2019facsimile} output 3D body meshes from images, but \emph{not clothing}.
This ignores image evidence that may be useful.
To reconstruct clothed bodies, methods use volumetric \cite{Natsume_2019_CVPR,saito2019pifu, varol2018bodynet,zheng2019deephuman} or bi-planar depth representations \cite{gabeur2910moulding} to model the body and garments as a whole.
We refer to these as Group 1 in Table~\ref{table:related_works}.
While these methods deal with arbitrary clothing topology and preserve a high level of detail, the reconstructed clothed body is not parametric, which means the pose, shape, and clothing of the reconstruction can not be controlled or animated.

Another group of methods are based on SMPL \cite{alldieck19cvpr,alldieck2018detailed, alldieck2018video, alldieck2019tex2shape,bhatnagar2019mgn, zhu2019detailed}. They represent clothing as an offset layer from the underlying body as proposed in ClothCap~\cite{pons2017clothcap}.
We refer to these approaches as Group 2 in Table~\ref{table:related_works}.
These methods can change the pose and shape of the reconstruction using the deformation model of SMPL.
This assumes clothing deforms like an undressed human body; i.e.~that clothing shape and wrinkles
\emph{do not change as a function of pose}.
We also use a body-to-cloth offset representation to learn our model, but critically, we learn a neural function mapping from pose to multi-modal clothing offset deformations.
Hence, our work differs from these methods in that we learn a parametric model of how clothing deforms with pose.

\myparagraph{Parametric models for 3D bodies and clothes.}
Statistical 3D human body models learned from 3D body scans, \cite{anguelov2005scape,joo2018total,loper2015smpl,pavlakos2019expressive} capture body shape and pose and are an important building block for multiple applications.
Most of the time, however, people are {\em dressed} and these models do not represent clothing.
In addition, clothes deform as we move, producing changing wrinkles at multiple spatial scales.
While clothing models learned from real data exist, few generalize to new poses.
For example, Neophytou and Hilton~\cite{Neophytou2014layered} learn a layered garment model on top of SCAPE~\cite{anguelov2005scape} from dynamic sequences, but generalization to novel poses is not demonstrated. Yang \etal~\cite{yang2018analyzing} train a neural network to \emph{regress} a PCA-based representation of clothing, but show generalization on the same sequence or on the same subject. L{\"a}hner \etal~\cite{lahner2018deepwrinkles} learn a garment-specific pose-deformation model by \emph{regressing} low-frequency PCA components and high frequency normal maps.
While the visual quality is good, the model is garment-specific and does not provide a solution for full-body clothing.
Similarly, Alldieck et al.~\cite{alldieck2019tex2shape} use displacement maps with a UV-parametrization to represent surface geometry, but the result is only static.
Wang \etal~\cite{wang2018learning} allow manipulation of clothing with sketches in a static pose.
The Adam model \cite{joo2018total} can be considered clothed but the shape is very smooth and not pose-dependent.
Clothing models have been learned from physics simulation of clothing~\cite{guan2012drape, gundogdu2019garnet,patel20vtailor,santesteban2019learning}, but visual fidelity
is limited by the quality of the simulations.
Furthermore, the above methods are \emph{regressors} that produce single point estimates.
In contrast, our model is \emph{generative}, which allows us to \emph{sample} clothing.

A conceptually different approach infers the parameters of a physical clothing model from 3D scan sequences~\cite{stoll2010video}.
This generalizes to novel poses, but the inference problem is difficult and, unlike our model, the resulting physics simulator is not differentiable with respect to the parameters.

\myparagraph{Generative models on 3D meshes.}
Our model predicts clothing displacements on the graph defined by the SMPL mesh using graph convolutions \cite{bruna2014spectral}.
There is an extensive recent literature on methods and applications of graph convolutions \cite{defferrard2016convolutional,kipf2016semi,litany2017deformable,ranjan2018generatingcoma,verma2018feastnet}.
Most relevant here, Ranjan \etal~\cite{ranjan2018generatingcoma} learn a convolutional autoencoder using graph convolutions \cite{defferrard2016convolutional}
with mesh down- and up-sampling layers \cite{garland1997surface}. Although it works well for faces,
the mesh sampling layer makes it difficult to capture the local details, which are key in clothing.
In our work, we capture local details by extending the PatchGAN \cite{isola2017pix2pix} architecture to 3D meshes.

\section{Additive Clothed Human Model}
To model clothed human bodies, we factorize them into two parts: the minimally-clothed body, and a clothing layer represented as displacements from the body.
This enables us to naturally extend SMPL to a class of clothing types by treating clothing as an additional additive shape term.
Since SMPL is in wide use, our goal is to extend it in a way that is consistent with current uses, making it effectively a ``drop in'' replacement for SMPL.
\subsection{Dressing SMPL} \label{sec:smpl-c}

SMPL~\cite{loper2015smpl} is a generative model of human bodies that factors the surface of the body into shape ($\beta$) and pose ($\theta$) parameters.
As shown in Fig.~\ref{fig:smpl} (a), (b), the architecture of SMPL starts with a triangulated template mesh, $\bar{T}$, in rest pose, defined by $N=6890$ vertices. Given shape and pose parameters ($\beta, \theta)$, 3D offsets are added to the template,  corresponding to shape-dependent deformations ($B_S(\beta)$) and pose dependent deformations ($B_P(\theta)$).
The resulting mesh is then \textit{posed} using the skinning function $W$. Formally:
\begin{eqnarray}
T(\beta, \theta) & = & \bar{T} + B_S(\beta) + B_P(\theta)\label{eq:SMPL2} \\
M(\beta,\theta) & = & W(T(\beta,\theta),J(\beta),\theta,\mathcal{W} ) \label{eq:SMPL1}
\end{eqnarray}
\noindent where the blend skinning function $W(\cdot)$ rotates the rest pose vertices $T$ around the 3D joints $J$ (computed from $\beta$), linearly smoothes them with the blend weights $\mathcal{W}$, and returns the posed vertices $M$.
The pose $\theta\in\mathbb{R}^{3\times(23+1)}$ is represented by a vector of relative 3D rotations of the $23$ joints and the global rotation in axis-angle representation.

SMPL adds {\it linear deformation layers} to an initial body shape. Following this, we define clothing as an extra offset layer from the body and add it on top of the SMPL mesh, Fig.~\ref{fig:smpl} (d).
In this work, we parametrize the clothing layer by the body pose $\theta$, clothing type $c$ and a low-dimensional latent variable $z$ that encodes clothing shape and structure.

Let $S_\textrm{clo}(z, \theta, c)$ be the clothing displacement layer. We extend Eq.~\eqref{eq:SMPL2} to a clothed body template in the rest pose:
\begin{equation}\label{eq:SMPL-clothing}
T_{\textrm{clo}}(\beta,\theta, c, z) = T(\beta,\theta) + S_\textrm{clo}(z, \theta, c) .
\end{equation}
Note that the clothing displacements, $S_\textrm{clo}(z, \theta, c)$, are pose-dependent.
The final clothed template is then posed with the SMPL skinning function, Eq.~\eqref{eq:SMPL1}:
\begin{eqnarray}
M(\beta,\theta,c, z) = W(T_{\textrm{clo}}(\beta,\theta, c, z),J(\beta),\theta,\mathcal{W} ). \label{eq:smplclothed}
\end{eqnarray}
This differs from simply applying blend skinning with \textit{fixed} displacements, as done in e.g.~\cite{alldieck19cvpr, bhatnagar2019mgn}.
Here, we train the model such that pose-dependent clothing displacements in the template pose are correct once posed by blend skinning.

\subsection{Clothing representation}
\label{sec:representation}
Vertex displacements are not a physical model for clothing and cannot represent all types of garments, but this approach
achieves a balance between expressiveness and simplicity, and has been widely used in deformation modeling~\cite{guan2012drape}, 3D clothing capture~\cite{pons2017clothcap} and recent work that reconstructs clothed humans from images~\cite{alldieck19cvpr,bhatnagar2019mgn,zhu2019detailed}.

The displacement layer is a graph $\mathcal{G}_d = (\mathcal{V}_d, \mathcal{E}_d)$ that inherits the SMPL topology: the edges $\mathcal{E}_d=\mathcal{E}_\textrm{SMPL}$.
$\mathcal{V}_d\in\mathbb{R}^{3\times N}$ is the set of vertices,
and the feature on each vertex is the 3-dimensional offset vector, $(d_x, d_y, d_z)$, from its corresponding vertex on the underlying body mesh.

We train our model on 3D scans of people in clothing.
From data pairs $(\mathcal{V}_\textrm{clothed},
\mathcal{V}_\textrm{minimal})$ we compute displacements, where $\mathcal{V}_\textrm{clothed}$ stands for the vertices of a clothed human mesh, and $\mathcal{V}_\textrm{minimal}$ the vertices of a minimally-clothed mesh.
Therefore, we first scan subjects in both clothed and minimally-clothed conditions, then use the SMPL model with free deformation~\cite{alldieck19cvpr,Zhang_2017_CVPR} to register the scans.
As a result, we obtain SMPL meshes capturing the geometry of the scans, the corresponding pose parameters, and vertices of the {\it unposed meshes}\footnote{We follow SMPL and use the T-pose as the zero-pose. For the mathematical details of registration and unposing, we refer the reader to~\cite{Zhang_2017_CVPR}.}. For each $(\mathcal{V}_\textrm{clothed}, \mathcal{V}_\textrm{minimal})$ pair, the displacements are then calculated as $\mathcal{V}_d = \mathcal{V}_{\textrm{clothed}} - \mathcal{V}_{\textrm{minimal}}$, where the subtraction is performed per-vertex along the feature dimension. Ideally, $\mathcal{V}_d$ has non-zero values only on body parts covered with clothes.

In summary, \app extends the SMPL body model to clothed bodies. Compared to volumetric representations of clothed people~\cite{Natsume_2019_CVPR,saito2019pifu,varol2018bodynet,zheng2019deephuman}, our combination of the body model and the garment layer is superior in the ease of re-posing and garment retargeting: the former uses the same blend skinning as the body model, while the latter is a simple addition of the displacements to a minimally-clothed body shape.
In contrast to similar models that also dress SMPL with offsets~\cite{alldieck19cvpr,bhatnagar2019mgn}, our garment layer is parametrized, low-dimensional, and pose-dependent.

\section{CAPE}
\begin{figure*}[!tb]
\centering
\begin{overpic}[width=\textwidth]{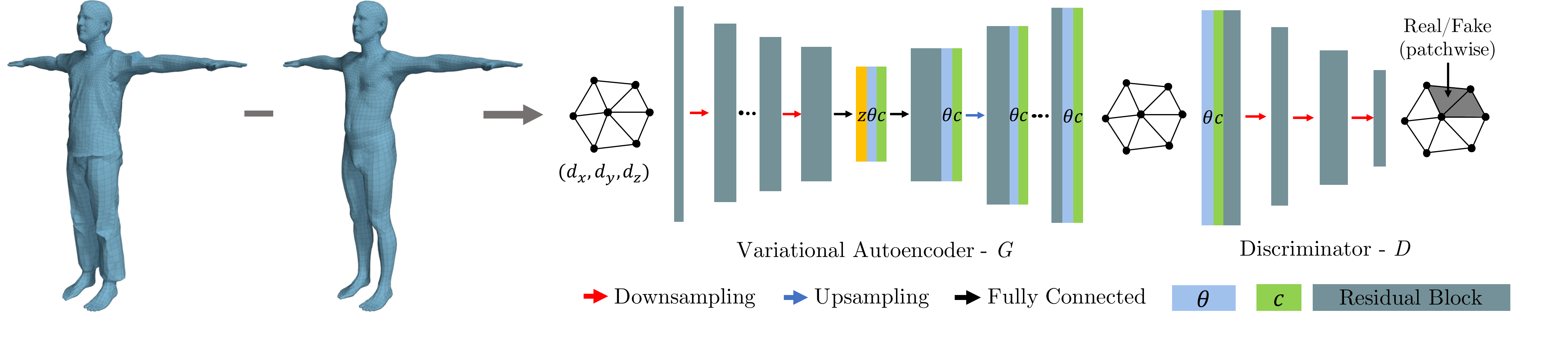}
\put(16.2, -0.5){\scriptsize \textbf{(a)}}
\put(70.5,-0.5){\scriptsize \textbf{(b)}}
\end{overpic}
\caption{Overview of our method. \textbf{(a)} Computing displacements
  from scan data (Sec.~\ref{sec:representation}) by subtracting the
  minimal body shape from clothed body mesh in the unposed space. \textbf{(b)} Schematic of our network architecture.}
\label{fig:cape_architecture}
\vspace{-0.4em}
\end{figure*}

\label{sec:cape}
Our clothing term $S_\textrm{clo}(z, \theta, c)$ in Eq.~\eqref{eq:SMPL-clothing} is a function of $z$, a code in a learned low-dimensional latent space that encodes the shape and structure of clothing, body pose $\theta$, and clothing type $c$. The function outputs the clothing displacement graph $\mathcal{G}_d$ as described in Sec.~\ref{sec:representation}.
We parametrize this function using a graph convolutional neural network (Graph-CNN) as a VAE-GAN framework~\cite{goodfellow2014gan,kingma2013auto,larsen2016autoencoding}.

\subsection{Network architecture}\label{sec:network}
As shown in Fig.~\ref{fig:cape_architecture}, our model consists of  a
generator $G$ with an encoder-decoder architecture and  a discriminator $D$. We also use auxiliary networks $C_1, C_2$ to handle the conditioning.
The network is differentiable and is trained end-to-end.

For simplicity, we use the following notation in this section. $x$: the vertices $\mathcal{V}_d$ of the input displacement graph; $\hat{x}
$: vertices of the reconstructed graph;
$\theta$ and $c$: the pose and clothing type condition vector; $z$: the latent code.

\myparagraph{Graph generator.}
We build the graph generator following a VAE-GAN framework. 
During training, an encoder $Enc(\cdot)$ takes in the displacement $x$, extract its features through multiple graph convolutional layers, and maps it to the low-dimensional latent code $z$.
A decoder is trained to reconstruct the input graph $\hat{x}\!=\! {Dec}(z)$ from $z$.
Both the encoder and decoder are feed-forward neural networks built with mesh convolutional layers.
Linear layers are used at the end of the encoder and the beginning of
the decoder. 
The architecture is shown in the supplemental materials.

Stacking graph convolution layers causes a loss of local features~\cite{li2018deeper} in the deeper layers.
This is undesirable for clothing generation because fine details, corresponding to wrinkles, are likely to disappear.
Therefore, we improve the standard graph convolution layers with residual connections, which enable the use of low-level features from the layer input if necessary.

At test time, the encoder is not needed. Instead, $z$ is sampled from the Gaussian prior distribution, and the decoder serves as the graph generator: $G(z)\! =\! Dec(z)$.
We detail different use cases below.

\myparagraph{Patchwise discriminator.}
To further enhance fine details in the reconstructions, we introduce a patchwise discriminator $D$ for graphs, which has shown success in the image domain~\cite{isola2017pix2pix,zhu2017unpaired}. Instead of looking at the entire generated graph, the discriminator only classifies whether a graph patch is real or fake based on its local structure. Intuitively this encourages the discriminator to only focus on fine details, and the global shape is taken care of by the reconstruction loss. 
We implement the graph patchwise-discriminator using four graph convolution-downsampling blocks~\cite{ranjan2018generatingcoma}. We add a discriminative real / fake loss for each of the output vertices. This enables the discriminator to capture a patch of neighboring nodes in the reconstructed graph and classify them as real / fake (see Fig.~\ref{fig:cape_architecture}).

\myparagraph{Conditional model.}
We condition the network with body pose $\theta$ and clothing type $c$.
The SMPL pose parameters are in axis-angle representation, and are difficult for the neural network to learn~\cite{kolotouros2019convolutional,lassner2017unite}. Therefore, following previous work~\cite{kolotouros2019convolutional,lassner2017unite}, we transform the pose parameters into rotational matrices using the Rodrigues equation.
The clothing types are discrete by nature, and we represent them using one-hot labels. 
Both conditions are first passed through a small fully-connected embedding network, $C_1(\theta), C_2(c)$, respectively, so as to balance the dimensionality of learned graph features and of the condition features.
We also experiment with different ways of conditioning the mesh generator: concatenation in the latent space; appending the condition features to the graph features at all nodes in the generator; and the combination of the two. We find that the combined strategy works better in terms of network capability and the effect of conditioning.

\subsection{Losses and learning}
For reconstruction, we use an L1 loss over the vertices of the mesh $x$, because it encourages less smoothing compared to L2, given by
\begin{equation}\label{eq:loss_recon}
\begin{split}
\mathcal{L}_{\textrm{recon}}  = \mathbb{E}_{x\sim p(x),z\sim q(z|x)}\left[\left\Vert G(z, \theta, c) - x \right\Vert_1\right].
\end{split}
\end{equation}

Furthermore, we apply a loss on the mesh edges to encourage the generation of wrinkles instead of smooth surfaces.
Let $e$ be an edge in the set of edges, $\mathcal{E}$, of the ground truth graph, and $\hat{e}$ the corresponding edge in the generated graph.
We penalize the mismatch of all corresponding edges by
\begin{equation}\label{eq:loss_edge}
\mathcal{L}_{\textrm{edge}} = \mathbb{E}_{e\in\mathcal{E},~\hat{e}\in\hat{\mathcal{E}}}\left[\left\Vert e - \hat{e} \right\Vert_2\right].
\end{equation}

We also apply a KL divergence loss between the distribution of latent codes and the Gaussian prior
\begin{equation}\label{eq:loss_kl}
\mathcal{L}_{\textrm{KL}} = \mathbb{E}_{x\sim p(x)}\left[\KLD{q(z|x)}{\mathcal{N}(0, \boldsymbol{I})}\right].
\end{equation}

Moreover, the generator and the discriminator are trained using an adversarial loss
\begin{equation}\label{eq:loss_gan}
\begin{split}
\mathcal{L}_{\textrm{GAN}} = ~&\mathbb{E}_{x\sim p(x)}\left[\log(D(x, \theta, c))\right] + \\
&\mathbb{E}_{z\sim q(z|x)}\left[\log(1-D(G(z,\theta, c)))\right],
\end{split}
\end{equation}
where $G$ tries to minimize this loss against the $D$ that aims to maximize it.

The overall objective is a weighted sum of these loss terms given by
\begin{equation}\label{eq:loss_total}
\mathcal{L} = \mathcal{L}_\textrm{recon} + \gamma_\textrm{edge}\mathcal{L}_\textrm{edge} + \gamma_\textrm{kl}\mathcal{L}_\textrm{KL} + \gamma_\textrm{gan}\mathcal{L}_\textrm{GAN}.
\end{equation}

\noindent Training details are provided in the supplemental materials. 

\section{CAPE Dataset}
\label{sec:dataset}
We build a dataset of 3D clothing by capturing temporal sequences of 3D human body scans with a high-resolution body scanner (3dMD LLC, Atlanta, GA).
Approximately 80K 3D scan frames are captured at 60 FPS, and a mesh with SMPL model topology is registered to each scan to obtain surface correspondences.
We also scanned the subjects in a minimally-clothed condition to obtain an accurate estimate of their body shape under clothing.
We extract the clothing as displacements from the minimally-clothed body as described in Sec.~\ref{sec:representation}.
Noisy frames and failed registrations are removed through manual inspection.

The dataset consists of 8 male subjects and 3 female subjects, performing a wide range of motions.
The subjects gave informed written consent to participate and to release the data for research purposes.
``Clothing type'' in this work refers to the 4 types of full-body outfits, namely \textit{shortlong}: short-sleeve upper body clothing and long lower body clothing; and similarly \textit{shortshort, longshort, longlong}. These outfits comprise 8 types of common garments. 
We refer to the supplementary material for the list of garments, further details, and examples from the dataset.

Compared to existing datasets of 3D clothed humans, our dataset provides captured data and \textit{alignments} of SMPL to the scans, separates the clothing from body, and provides accurate, \textit{captured} ground truth body shape under clothing.
For each subject and outfit, our dataset contains large pose variations, which induces a wide variety of wrinkle patterns.
Since our dataset of 3D meshes has a consistent topology, it can be used for the quantitative evaluation of different Graph-CNN architectures. The dataset is available for research purposes at \websiteCAPE.

\section{Experiments}
We first show the representation capability of our model
and then demonstrate the model's ability to generate new examples by probabilistic sampling.
We then show an application to human pose and shape estimation.

\subsection{Representation power}\label{sec:recon}
\myparagraph{3D mesh auto-encoding errors.}
We use the reconstruction accuracy to measure the capability of our VAE-based model for geometry encoding and preserving.
We compare with a recent convolutional mesh autoencoder, CoMA \cite{ranjan2018generatingcoma}, and a linear (PCA) model.
We compare to both the original CoMA with a 4$\times$ downsampling (denoted as ``CoMA-4''), and without downsampling (denoted  ``CoMA-1'') to study the effect of downsampling on over-smoothing.
We use the same latent space dimension $n_z=18$ (number of principal components in the case of PCA) and hyper-parameter settings, where applicable, for all models.

Table \ref{table:recon_compare} shows the  per-vertex Euclidean error when using our network to reconstruct the clothing displacement graphs from a held-out test set in our CAPE dataset. The model is trained and evaluated on male and female data separately.
Body parts such as head, fingers, toes, hands and feet are excluded
from the accuracy computation, as they are not covered with clothing. 

Our model outperforms the baselines in the auto-encoding task; additionally, the reconstructed shape from our model is \textit{probabilistic} and \textit{pose-dependent}.
Note that, CoMA here is a deterministic auto-encoder with a focus on reconstruction.
Although the reconstruction performance of PCA is on par with our
method on male data, PCA can not be used directly in the inference
phase with a pose parameter as input. 
Furthermore, PCA assumes a Gaussian distribution of the data, which
does not hold for complex clothing deformations. 
Our method addresses both of these issues.

\begin{figure}[!bt]
\begin{center}
   \includegraphics[width=\linewidth]{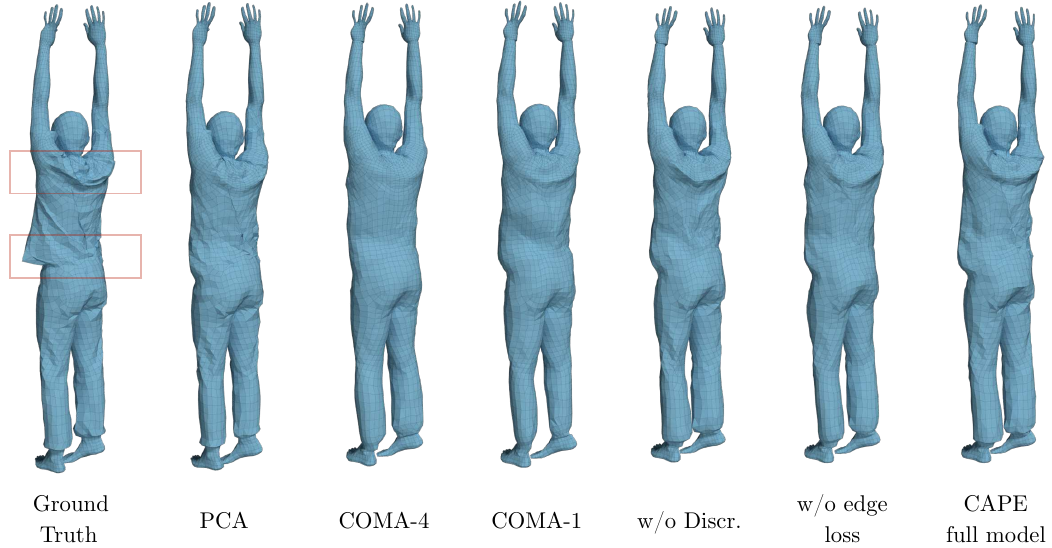}
\end{center}
\vspace{-0.5em}
\caption{Example of reconstruction by the baselines, ablated version of our model, and our full model. Pay attention to the hem and wrinkles on upper back. Our model is able to recover both long-range correlations and local details. Best viewed zoomed-in on screen.}
\label{fig:qualitative_recon}
\end{figure}

Fig.~\ref{fig:qualitative_recon} shows a qualitative comparison of the methods.
PCA keeps wrinkles and boundaries, but the rising hem on the left side
disappears. CoMA-1 and CoMA-4 are able to capture global correlation,
but the wrinkles tend to be smoothed. By incorporating all the key
components, our model manages to model both local structures and
global correlations more accurately than the other methods.

\begin{table}[b]
\scriptsize
\centering
\vspace{2pt}
\caption{Per-vertex auto-encoding error in millimeters. Upper section: comparison with baselines; lower section: ablation study.}
\label{table:recon_compare}
\begin{tabular}{l|cc|cc}
\hline
& \multicolumn{2}{c|}{Male} & \multicolumn{2}{c}{Female} \\
\hline
{\bf Methods} & {Error mean} & {median} & {Error mean} & {median} \\
PCA& 5.65\tpm\textbf{4.81} & 4.30 & 4.82\tpm3.82 & 3.78 \\
CoMA-1 & 6.23\tpm5.45 & 4.66 & 4.69\tpm3.85 & 3.61 \\
CoMA-4 & 6.87\tpm5.62 & 5.29 & 4.86\tpm3.96 & 3.75 \\ 
{\bf Ours} & \textbf{5.54}\tpm5.09 & \textbf{4.03} & \textbf{4.21}\tpm\textbf{3.76} & \textbf{3.08}\\
\hline
{\bf Ablated Components} & {Error mean} & {median} & {Error mean} & {median} \\
Discriminator & 5.65\tpm5.18& 4.11 & 4.31\tpm3.78& 3.18\\
Res-block & 5.60\tpm5.21& 4.05 & 4.27\tpm3.76& 3.15\\
Edge loss & 5.93\tpm5.40& 4.32 & 4.32\tpm3.78& 3.19\\
\hline
\end{tabular}
\end{table}

\myparagraph{Ablation study.}
We remove key components from our model while keeping all the others, and evaluate the model performance; see  Table \ref{table:recon_compare}. We observe that the discriminator, residual block and edge loss all play important roles in the model performance.
Comparing the performance of CoMA-4 and CoMA-1, we find that the mesh the downsampling layer causes a loss of fidelity.
However, even without any spatial downsampling, CoMA-1 still
underperforms our model. This shows the benefits of adding the discriminator, residual block, and edge loss in our model.

\subsection{Conditional generation of clothing}\label{sec:sample_experients}

\begin{figure*}[tb]
\begin{center}
   \includegraphics[width=\textwidth]{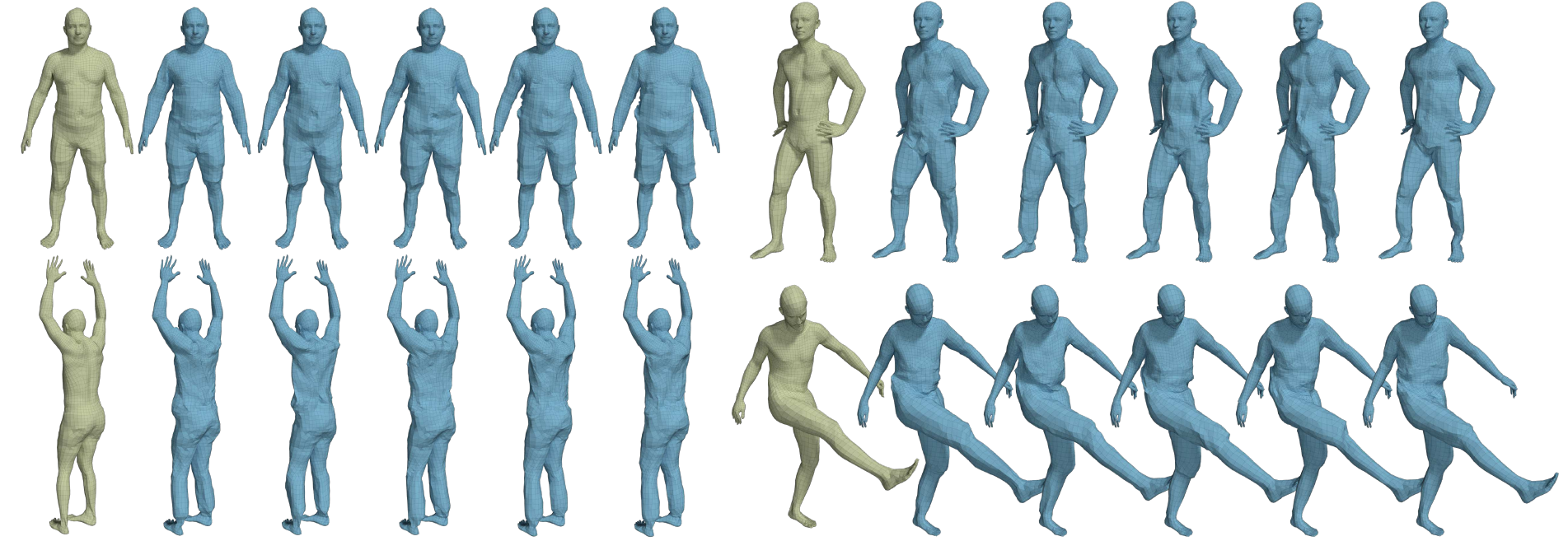}
\end{center}
\vspace{-0.1in}
\caption{Clothing sampled and generated from our \app model applied to
  four unseen body shapes (green colored) with different poses. 
Each body in green is followed by five blue examples (generated by
sampling the latent vector $z$) showing different versions of the same outfit type. The four groups are wearing outfit types ``longshort'', ``longlong'', ``shortlong'' and ``shortshort'', respectively. Best viewed zoomed-in on screen.}
\label{fig:sample_experiment}
\vspace{-0.2em}
\end{figure*}

As a generative model, \app can be sampled and generates new data.
The model has three parameters: $z, c, \theta$ (see Eq.~\eqref{eq:SMPL-clothing}). By sampling one of them while keeping the other two fixed, we show how the conditioning affects the generated clothing shape.

\myparagraph{Sampling.}
Fig.~\ref{fig:sample_experiment} presents the sampled clothing dressed on unseen bodies, in a variety of poses that are not used in training.
For each subject, we fix the pose $\theta$ and clothing type $c$, and sample $z$ several times to generate varied clothing shapes.
The sampling trick in \cite{ghosh2019variational} is used. Here we only show untextured rendering to highlight the variation in the generated \textit{geometry}.
As \app inherits the SMPL topology, the generated clothed body meshes are compatible with all existing SMPL texture maps. 
See supplemental materials for a comparison between a CAPE sample and
a SMPL sample rendered with the same texture.

As shown in the figure, our model manages to capture long-range
correlations within a mesh, such as the elevated hem for a subject with raised arms, and the lateral wrinkle on the back with raised arms.
The model also synthesizes local details such as wrinkles in the armpit area, and boundaries at cuffs and collars.

\begin{figure}[htb]
\begin{overpic}[width=0.9\linewidth]{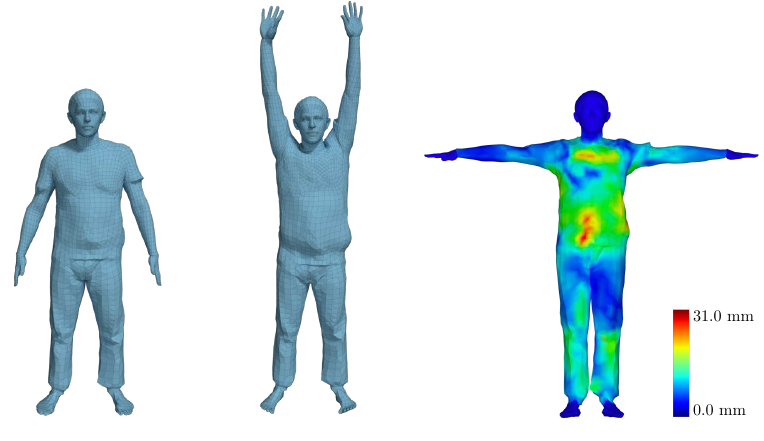}
\put(10, -0.5){\scriptsize \textbf{(a)}}
\put(39.1, -0.5){\scriptsize \textbf{(b)}}
\put(76.2, -0.5){\scriptsize \textbf{(c)}}
\end{overpic}
\vspace{0.5em}
\caption{Pose-dependent clothing shape. \textbf{(a)} and \textbf{(b)}: two clothing shapes generated from \app, with different pose parameters. \textbf{(c)}: color-coded difference of the offset clothing layers in \textbf{(a)} and \textbf{(b)}, in the canonical pose.}\label{fig:pose-dependent}
\end{figure}

\myparagraph{Pose-dependent clothing deformation.}
Another practical use case of \app is to animate an existing clothed
body. This corresponds to fixing the clothing shape variable $z$ and
clothing type $c$, and reposing the body by changing $\theta$. The
challenge here is to have a clothing shape that is consistent across
poses, yet deforms plausibly. We demonstrate the pose-dependent effect
on a test pose in Fig.~\ref{fig:pose-dependent}. The difference of the
clothing layer between the two poses is calculated in the canonical
pose, and shown with color coding. The result shows that the clothing type is consistent while local deformation changes along with pose.
We refer to the supplemental video for a comparison with traditional rig-and-skinning methods that use fixed clothing offsets.

\myparagraph{User study of generated examples.}
To test the realism of the generated results from our method, we performed a user study on Amazon Mechanical Turk (AMT).
We dress virtual avatars in 3D and render them into front-view images. Following the protocol from \cite{isola2017pix2pix}, raters are presented with a series of ``real vs fake'' trials. On each trial, the rater is presented with a ``real'' mesh render (randomly picked from our dataset), 
and a ``fake'' render (generated by our model), shown side-by-side. The raters are asked to pick the one that they think is real. Each pair of renderings is evaluated by 10 raters. More strictly than \cite{isola2017pix2pix}, we present both real and fake renderings \textit{simultaneously}, do not set a time limit for raters and allow zoom-in for detailed comparison. In this setting, the best score that a method can obtain is 50\%, meaning that the real and fake examples are indistinguishable.

We evaluate with two test cases. In test case 1, we fix the clothing type to be ``shortlong'' (the most common clothing type in training), and generate 300 clothed body meshes with various poses for the evaluation.
In test case 2, we fix the pose to be an A-pose (the most frequent pose in training), and sample 100 examples per clothing type for evaluation. 
On average, in the direct comparison with real data, our synthesized
data ``fools'' participants $35.1\%\pm15.7\%$ and $38.7\%\pm16.5\%$ of
the time respectively (\ie these participants labeled our result as
``real'').

\subsection{Image fitting}\label{sec:image_fitting_experiment}
\app is fully differentiable with respect to the clothing shape
variable $z$, body pose $\theta$ and clothing type $c$. Therefore, it
can also be used in optimization frameworks. We show an application of
\app on the task of reconstructing a body mesh from a single image, by
entending the optimization-based method,
{SMPLify}~\cite{bogo2016smplify}.
Assuming $c$ is known, 
we dress the minimally-clothed output mesh from SMPLify using \app, project it back to the image using a differentiable renderer \cite{henderson19ijcv} and optimize for $\beta, \theta, z$, with respect to the silhouette discrepancy.

\begin{table}[b]
\scriptsize
\centering
\caption{Vertex MSE of image fitting results measured in meters.}%
\label{table:img_fitting_exp}
\begin{tabular}{lcc}
\hline
Method & SMPLify~\cite{bogo2016smplify} & Ours\\
\hline%
Per-vertex MSE & 0.0223 & 0.0189\\
\hline
\end{tabular}
\end{table}
We evaluate our image fitting pipeline on renderings of 120 randomly selected unseen test examples from the \app dataset. To compare, we measure the reconstruction error of SMPLify and our results against ground truth meshes using mean square vertex error (MSE). To eliminate the error introduced by the ambiguity of human scale and distance to the camera, we optimize the global scaling and translation of predictions for both methods on each test sample. 
A mask is applied to exclude error in the non-clothed regions such as head, hands and feet. We report the errors of both methods in Table~\ref{table:img_fitting_exp}.
Our model performs 18\% better than SMPLify due to its ability to capture clothing shape.

More details about the objective function, experimental setup and qualitative results of the image fitting experiment are provided in supplementary materials.

Furthermore, once a clothed human is reconstructed from the image, our
model can repose and \textit{animate} it, as well as change the
subject's clothes by re-sampling $z$ or clothing type $c$. This shows
the potential for several applications. We show examples in the supplemental video.

\section{Conclusions, Limitations, Future Work}
We have introduced a novel graph-CNN-based generative shape model that enables us to condition, sample, and preserve fine shape detail in 3D meshes.
We use this to model clothing deformations from a 3D body mesh and condition the latent space on body pose and clothing type.
The training data represents 3D displacements from the SMPL body model for varied clothing and poses.
This design means that our generative model is compatible with SMPL in that clothing is an additional {\em additive} term applied to the SMPL template mesh.
This makes it possible to sample clothing, dress SMPL with it, and then animate the body with pose-dependent clothing wrinkles.
A clothed version of SMPL has wide applicability in computer vision.
As shown, we can apply it to fitting the body to images of clothed humans.
Another application would use the model to
generate training data of 3D clothed people to train regression-based pose-estimation methods.

There are a few limitations of our approach that point to future work.
First, CAPE inherits the limitation of the offset representation for
clothing: (1) Garments such as skirts and open jackets differ from the body topology and cannot be trivially represented by offsets. 
Consequently, when fitting CAPE to images containing such garments, it
could fail to explain the image evidence; see discussions on the skirt
example in the supplementary material.
(2) Mittens and shoes: they can technically be modeled by the offsets, but their geometry is sufficiently different from fingers and toes, making this impractical.
A multi-layer model can potentially overcome these limitations.
Second, the level of geometric details that CAPE can achieve is upper-bounded by the mesh resolution of SMPL. To produce finer wrinkles, one can resort to higher resolution meshes or bump maps. 
Third, while our generated clothing depends on pose, it does not depend on {\em dynamics}.
This does not cause severe problem for most slow motions but does not generalize to faster motions.
Future work will address modeling clothing deformation on temporal sequence and dynamics.

{\footnotesize
\myparagraph{Acknowledgements:}
We thank Daniel~Scharstein for revisions of the manuscript, Joachim~Tesch for the help with Blender rendering, Vassilis Choutas for the help with image fitting experiments, and Pavel~Karasik for the help with AMT evaluation. We thank Tsvetelina~Alexiadis and Andrea~Keller for collecting data. We thank Partha~Ghosh, Timo~Bolkart and Yan~Zhang for useful discussions. Q.~Ma and S.~Tang acknowledge funding by Deutsche Forschungsgemeinschaft (DFG, German Research Foundation) - 276693517 SFB 1233. G.~Pons-Moll is funded by the Emmy Noether Programme, Deutsche Forschungsgemeinschaft - 409792180.

\myparagraph{Disclosure:~}MJB has received research gift funds from Intel, Nvidia, Adobe, Facebook, and Amazon. 
While MJB is a part-time employee of Amazon, his research was performed solely at, and funded solely by, MPI. 

\bibliographystyle{ieee_fullname}
\bibliography{references}
}

\clearpage
\appendix
{\noindent\Large\textbf{Supplementary Material}}
\setcounter{page}{1}

\section{Implementation Details}
\subsection{CAPE network architecture}\label{appendix:architecture}
Here we provide the details of the CAPE architecture, as discribed in the main paper Sec.~\ref{sec:network}.
\noindent We use the following notations:
\begin{itemize}
\setlength\itemsep{0em}
\item $x$: data, $\hat{x}$: output (reconstruction) from the decoder, $z$: latent code, $p$: the prediction map from the discriminator.
\item LReLU: leaky rectified linear units with a slope of $0.1$ for negative values.
\item $\textrm{CONV}_{n}$: Chebyshev graph convolution layer with $n$ filters.
\item $\textrm{CONVBlock}_{n}$: convolution block comprising $\textrm{CONV}_{n}$ and LReLU.
\item $\textrm{CResBlock}_n$: conditional residual block that uses $\textrm{CONV}_{n}$ as filters.
\item $\textrm{DS}_{n}$: linear graph downsampling layer with a spatial downsample rate $n$.
\item $\textrm{US}_{n}$: linear graph upsampling layer with a spatial upsample rate $n$.
\item $\textrm{FC}_{m}$: fully connected layer with output dimension $m$.
\end{itemize}

\myparagraph{Condition module:}
for pose $\theta$, we remove the parameters that are not related to clothing, \eg head, hands, fingers, feet and toes, resulting in 14 valid joints from the body. The pose parameters from each joint are represented by the flattened rotational matrix (see Sec.~\ref{sec:network}, ``Conditional model''). This results in the 
overall pose parameter $\mathbb{R}^{9\times14}$. We feed this into a small fully-connected network:
\begin{align*}
\theta \in \mathbb{R}^{9\times14} &\to \fc{63} \to\relu&\\
               &\to\fc{24} \to z_\theta \in \mathbb{R}^{24}
\end{align*}

\noindent The clothing type $c$ refers to the type of ``outfit'', \ie a combination of upper body clothing and lower body clothing. There are four types of outfits in our training data: \textit{longlong}: long sleeve shirt / T-shirt / jersey with long pants; \textit{shortlong}: short sleeve shirt / T-shirt / jersey with long pants; and their opposites, \textit{shortshort} and \textit{longshort}. As the types of clothing are discrete by nature, we represent them using a one-hot vector, $c \in \mathbb{R}^{4}$, and feed it into a linear layer:
\begin{align*}
c \in \mathbb{R}^{4}\to \fc{8} \to z_c \in \mathbb{R}^{8}
\end{align*}

\myparagraph{Encoder}:
\begin{align*}
x \in\mathbb{R}^{3\times6890} &\to \convblock{64} \to \ds \\
                &\to \convblock{64} \to \convblock{128} \to \ds  \\
                &\to \convblock{128} \to \convblock{256} \to \ds \\
                &\to \convblock{256} \to \convblock{512} \\
                &\to \convblock{512} \to \conv{64}^{1\times1} \\ 
                & \to \fc{18} \to z_\mu \in\mathbb{R}^{18} \\
                & \; \; \rotatebox[origin=c]{180}{$\Lsh$} \; \; \fc{18} \to z_\sigma\in\mathbb{R}^{18}
\end{align*}

\myparagraph{Decoder}:
\begin{align*}
z \in \mathbb{R}^{18} &\xrightarrow[z_\theta, z_c]{\text{concat}} z' \in \mathbb{R}^{18+24+8}\\
            &\to \fc{64\times862} \to  \conv{512}^{1\times1}\\
              &\to \cres{512} \to \cres{512} \to \us \\
              &\to \cres{256} \to \cres{256} \to \us \\
                &\to \cres{128} \to \cres{128} \to \us \\
                &\to \cres{64} \; \xspace \to \cres{64} \\
              &\to \textrm{CONV}_3 \to \hat{x} \in \mathbb{R}^{3\times6890}
\end{align*}

\myparagraph{Discriminator}:
\begin{align*}
\hat{x} \in\mathbb{R}^{3\times6890} &\xrightarrow[\textrm{tile}\{z_\theta, z_c\}]{\text{concat}} \hat{x}' \in \mathbb{R}^{(3+24+8)\times6890}\\
               & \to \convblock{64} \to \ds \\
               & \to \convblock{64} \to \ds \\
               & \to \convblock{128} \to \ds \\
               & \to \convblock{128} \to \ds \\
           & \to \textrm{CONV}_1 \to p \in \mathbb{R}^{1\times431}
\end{align*}

\myparagraph{Conditional residual block:}
We adopt the graph residual block from Kolotouros \etal~\cite{kolotouros2019convolutional} that includes Group Normalization \cite{wu2018group}, non-linearity, graph convolutional layer and graph linear layer (\ie Chebyshev convolution with polynomial order of $0$). After the input to the residual block, we append the condition vector to every input node along the feature channel. Our CResBlock is given by
\begin{align*}
x_\textrm{in} \in \mathbb{R}^{i\times P} & \xrightarrow[\textrm{tile}\{z_\theta, z_c\}]{\text{concat}} {x}' \in \mathbb{R}^{(i+24+8)\times P} \\
               & \to \res{j} \to x_\textrm{out} \in \mathbb{R}^{j\times P}
\end{align*}
where $x_\textrm{in}$ is the input to the $\cres{}$. $x_\textrm{in}$ has $P$ nodes and $i$ features on each node. ResBlock is the graph residual block from \cite{kolotouros2019convolutional} that outputs $j$ features on each node.

\subsection{Training details}\label{appendix:training_details}
The model is trained for 60 epochs, with a batch size of 16, using stochastic gradient descent with a momentum of $0.9$.
The learning rate starts from an initial value of $2\times10^{-3}$, increases with a warm-up step of $2\times10^{-3}$ / epoch for 4 epochs, and then decays with a rate of $0.99$ after every epoch.

The convolutions use the Chebyshev polynomial of order $2$ for the generator, and of order $3$ for the discriminator. An L2-weight decay with strength $2\times10^{-3}$ is used as regularization. 

We train and test our model for males and females separately. We split the male
dataset into a training set of 26,574 examples and 5,852 test
examples. The female dataset is split into a training set of 41,472
examples and a test set of 12,656 examples. Training takes
approximately 15 minutes per epoch on the male dataset and 20 minutes
per epoch on the female dataset.

\section{Image Fitting}\label{appendix:image_fitting}
Here we detail the objective function, experimental setup and extended results of the image fitting experiments, as described in the main manuscript Sec.~\ref{sec:image_fitting_experiment}.

\begin{figure*}[!t]
\begin{center}
   \includegraphics[width=0.95\linewidth]{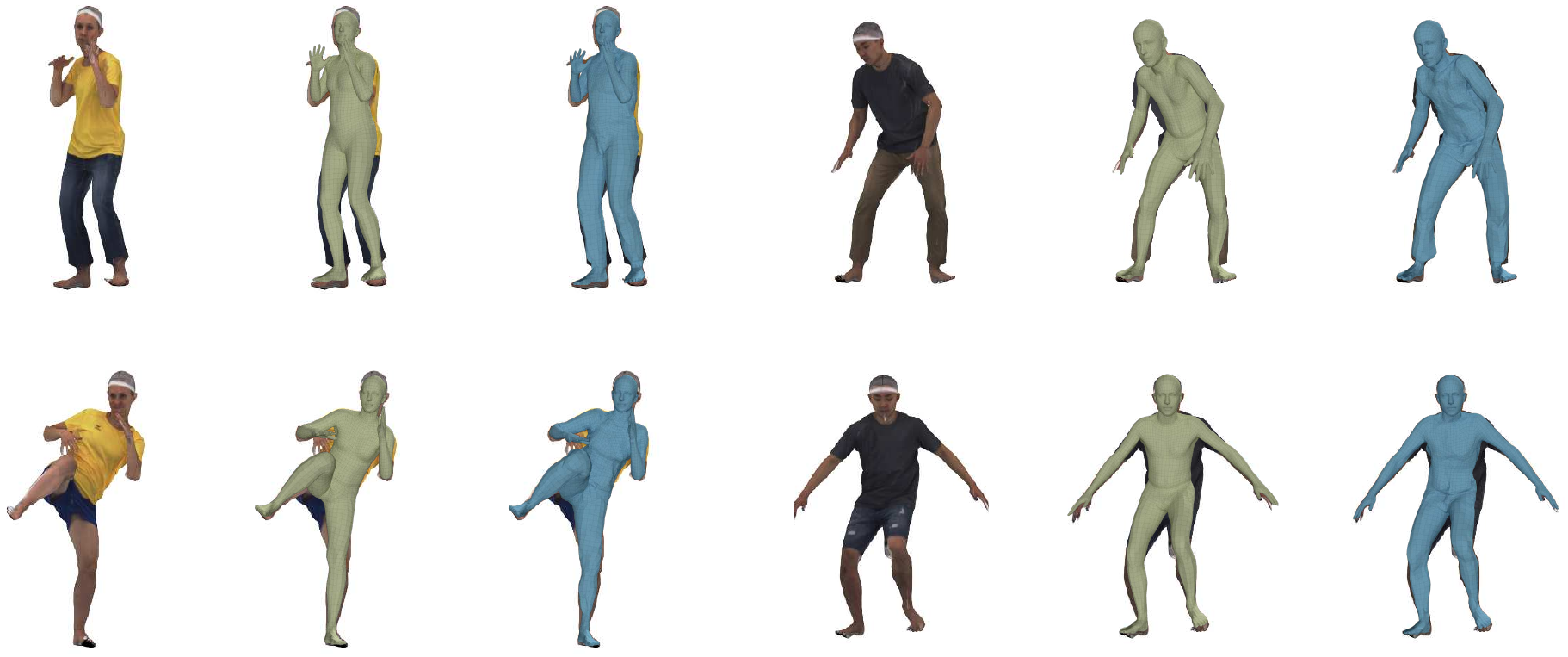}
\end{center}
\vspace{-1.2em}
\caption{Qualitative results on the rendered meshes from CAPE dataset. Minimally-clothed fitting results from SMPLify \cite{bogo2016smplify} are shown in green; results from our method are shown in blue.}
\label{fig:img_fitting_qualitative}
\end{figure*}

\subsection{Objective function}
Similar to \cite{lassner2017unite}, we introduce a silhouette term to encourage the shape of the clothed body to match the image evidence. The silhouette is the set of all pixels that belong to a body's projection onto the image. 
Let $\hat{S}(\beta, \theta, c, z)$ be the rendered silhouette of a clothed body mesh $M(\beta, \theta, c, z)$ (
see main paper Eq.~\eqref{eq:smplclothed}),
and $S$ be the ground truth silhouette. The silhouette objective is defined by the bi-directional distance between $S$ and $\hat{S}(\cdot)$:
\begin{align}\label{eq:silh_loss}\nonumber
E_S(\beta, \theta, z; c, S, K) & = \sum_{x\in\hat{S}}l(x, S) \\
                 & + \sum_{x\in S}l(x, \hat{S}(\beta, \theta, c, z))
\end{align}
where $l(x, S)$ is the L1 distance from a point x to the closest point in the silhouette $S$. The distance is zero if the point is inside $S$. $K$ is the camera parameter that is used to render the mesh to the silhouette on the image plane. The clothing type is derived from upstream pipeline and is therefore not optimized.

For our rendered scan data, the ground truth silhouette and clothing type are acquired for free during rendering. For in-the-wild images, this information can be acquired using human-parsing networks, \eg \cite{gong2018instance}.

After the standard SMPLify optimization pipeline, we apply the clothing layer to the body, and apply an additional optimization step on body shape $\beta$, pose $\theta$ and clothing structure $z$, with respect to the overall objective:
\begin{align}\label{eq:total_obj}
\nonumber E_\textrm{total} &=  E_J(\beta, \theta; K, J_\textrm{est}) + \lambda_S E_S(\beta, \theta, z; c, S, K) \\
\nonumber& + \lambda_\theta E_\theta(\theta) + \lambda_\beta E_\beta(\beta)\\
& + \lambda_a E_a(\theta) + \lambda_z E_z(z)
\end{align}
The overall objective is a weighted sum of the silhouette loss with other standard SMPLify energy terms.
$E_J$ is a weighted 2D distance between the projected SMPL joints and the detected 2D points, $J_\textrm{est}$. $E_\theta(\theta)$ is the mixture of Gaussians pose prior term, $E_\beta(\beta)$ the shape prior term, $E_a(\theta)$ the penalty term that discourages unnatrual joint bents, and $E_z(z)$ the L2-regularizer on $z$ to prevent extreme clothing deformations.
For more details about these terms please refer to Bogo \etal~\cite{bogo2016smplify}.

\subsection{Data} We render 120 textured meshes (aligned to the SMPL
topology) randomly selected from the test set of the CAPE dataset that include variations in gender, pose and clothing type, at a resolution of $512\times512$. The ground truth meshes are used for evaluation. Examples of the rendering are shown in Fig.~\ref{fig:img_fitting_qualitative}.

\subsection{Setup} We re-implement the SMPLify work by Bogo \etal~\cite{bogo2016smplify} in Tensorflow, using the gender neutral SMPL body model. Compared to the original SMPLify, there are two major changes. First, we do not include the interpenetration error term, as it slows down the fitting but brings little performance gain \cite{SPIN:ICCV:2019}. Second, we use OpenPose for the ground truth 2D keypoint detection instead of DeepCut \cite{pishchulin2016deepcut}.

\subsection{Evaluation} We measure the mean square error (MSE) between ground truth vertices $\mathcal{V}_\textrm{GT}$ and reconstructed vertices from SMPLify $\mathcal{V}_\textrm{SMPLify}$, and from our pipeline (Eq.~\eqref{eq:total_obj}) $\mathcal{V}_\textrm{CAPE}$, respectively. As discussed in Sec.~\ref{sec:image_fitting_experiment},
 to eliminate the influence of the ambiguity caused by focal length, camera translation and body scale, we estimate the body scale $s$ and camera translation $t$ for both $\mathcal{V}_\textrm{SMPLify}$ and $\mathcal{V}_\textrm{CAPE}$. Specifically, we optimize the following energy function for $\mathcal{V}=\mathcal{V}_\textrm{SMPLify}$ and $\mathcal{V}=\mathcal{V}_\textrm{CAPE}$ respectively:
\begin{equation}
 E = \argmin_{s,t} \frac{1}{N} \sum_{i \in \mathbf{C}}|| s(\mathcal{V}_i+t)- \mathcal{V}_{\textrm{GT},i} ||^2
\end{equation}
where $i$ is vertex index, $\mathbf{C}$ the set of clothing vertex indices, and $N$ the number of elements in $\mathbf{C}$. Then, the MSE is computed with estimated scale $\hat{s}$ and translation $\hat{t}$ using:
\begin{equation}
 \textrm{MSE} = \frac{1}{N} \sum_{i \in \mathbf{C}}||\hat{s}(\mathcal{V}_i+\hat{t}) - \mathcal{V}_{\textrm{GT},i}||^2
\end{equation}

\subsection{Extended image fitting results}
\begin{figure*}[!htb]
\begin{center}
   \includegraphics[width=0.98\textwidth]{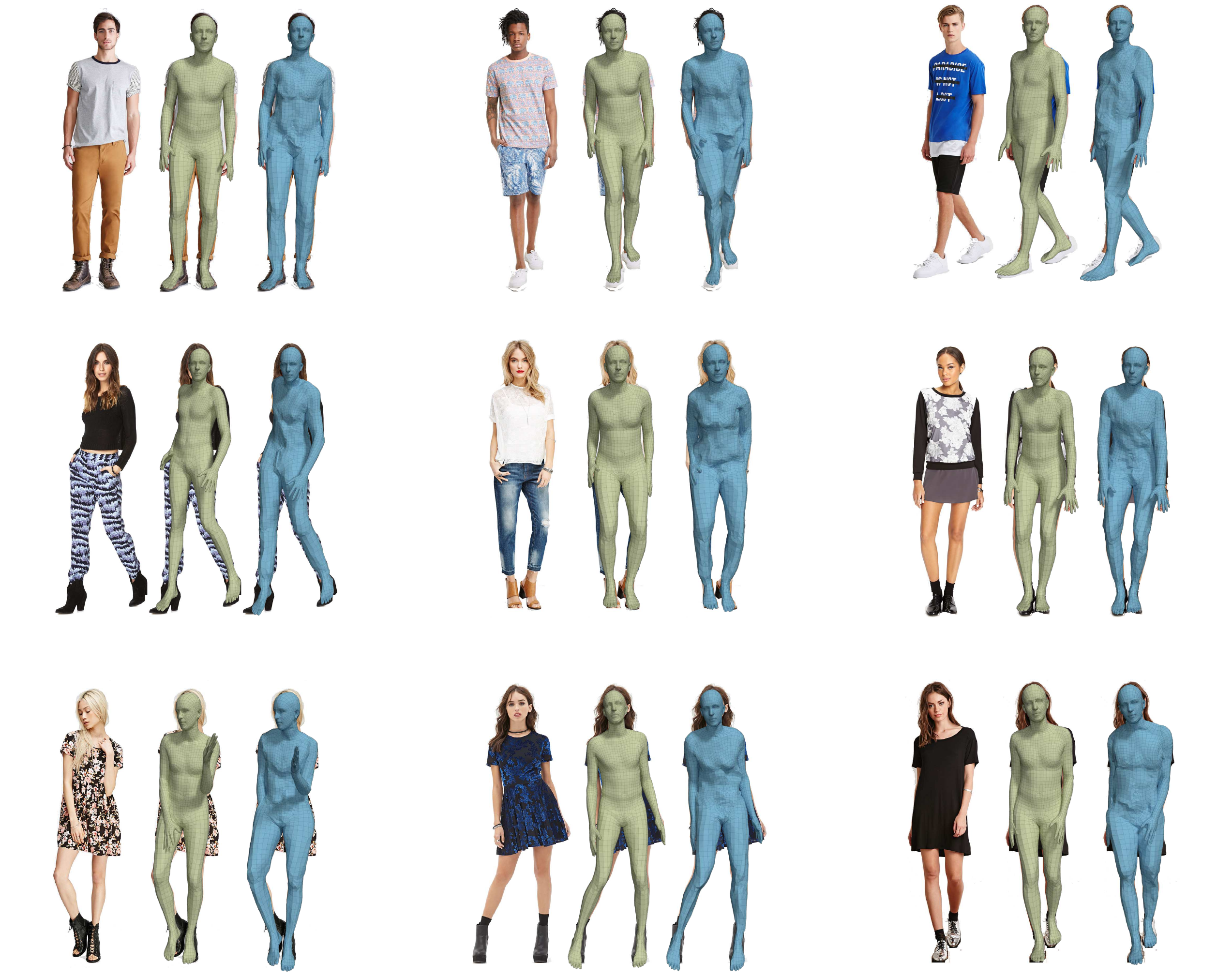}
\end{center}
\vspace{-1.2em}
\caption{Qualitative results on fashion images from the DeepFashion dataset~\cite{liu2016deepfashion}. SMPLify \cite{bogo2016smplify} results are shown in green, our results are in blue.}
\label{fig:deepfashion}
\end{figure*}

\myparagraph{Qualitative results.}
Fig.~\ref{fig:img_fitting_qualitative} shows the reconstruction result of SMPLify~\cite{bogo2016smplify} and our method on rendered meshes from the CAPE dataset. Quantitative results can be found in the main manuscript, Table~3. We also show qualitative results of CAPE fitted to images from the DeepFashion~\cite{liu2016deepfashion} dataset in Fig.~\ref{fig:deepfashion}. In general CAPE has better silhouette overlapping and in some cases improved pose estimation, but has also shown a few limitations that point to future work, as disucussed next.

\myparagraph{Limitations and failure cases.}
Since our image fitting pipeline is based on SMPLify, it fails when SMPLify fails to predict the correct pose. 
Besides, while in this work the reconstructed clothing geometry only relies on the silhouette loss, it can further benefit from other losses such as the photometric loss.
Recent regression-based methods have achieved improved performance on this task \cite{kanazawa2018end,SPIN:ICCV:2019,kolotouros2019convolutional}, and integrating CAPE with them is an interesting future line of work.

The CAPE model itself fails when the garment in the image is beyond its model space. As discussed in the main paper, CAPE inherits the limitations of the offset representation in terms of clothing types. Skirts, for example, have a different topology from human bodies, and can hence not be modeled by CAPE. Consequently, if CAPE is employed to fit images of people in skirts, it can only approximate with \eg the outfit type \textit{shortshort}, which fails to explain the observation in the image. The last row of Fig.~\ref{fig:deepfashion} shows a few of such failure cases on skirt images from the DeepFashion dataset~\cite{liu2016deepfashion}. 
Despite better silhouette matching than the minimally-clothed fitting, the reconstructed clothed bodies have the wrong garment type, which do not match the evidence in the image. Future work can explore multi-layer clothing models that can handle these garment types.

\subsection{Post-image fitting: re-dress and animate}
After reconstructing the clothed body from the image, CAPE is capable of dressing the body with new styles by sampling the $z$ variable, changing the clothing type by sampling $c$, and animating the mesh by sampling the pose parameter $\theta$. We provide such a demo in the supplemental video\footnote{ available at \websiteCAPE.}.
This shows the potential in a wide range of applications.

\section{Extended Experimental Results}
\subsection{CAPE with SMPL texture}\label{appendix:cape_with_texture}
\begin{figure}[!hb]
\begin{center}
   \includegraphics[width=0.98\linewidth]{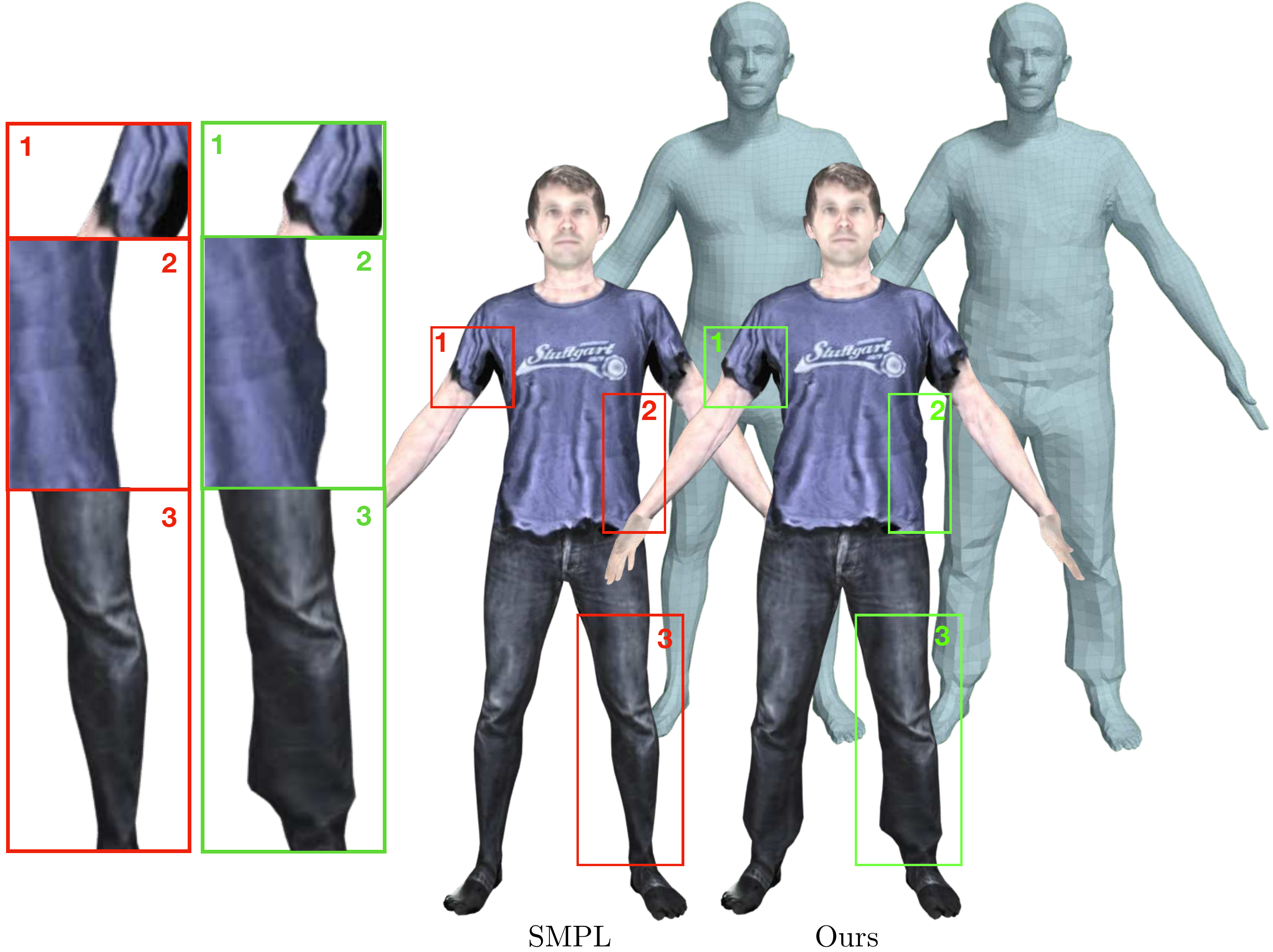}
\end{center}
\vspace{-1.2em}
\caption{Front row: A clothing texture applied to the SMPL body and one of our generated clothed bodies. Back row: respective underlying geometry. Best viewed zoomed-in on screen.}
\label{fig:textured}
\end{figure}

As our model has the same topology as SMPL, it is compatible with all existing SMPL texture maps, which are mostly of clothed bodies. Fig.~\ref{fig:textured}
shows an example texture applied to  the standard minimally-clothed SMPL model (as done in the SURREAL dataset \cite{varol2017learning}) and to our clothed body model, respectively. Although the texture creates an illusion of clothing on the SMPL body, the overall shape remains skinny, oversmoothed, and hence unrealistic. In contrast, our model, with its improved clothing geometry, matches more naturally the clothing texture if the correct clothing type is given.
This visual contrast becomes even stronger when the texture map has no shading information (albedo map), and when the object is viewed in a 3D setting. See \texttt{03:02} in the supplemental video for the comparison in 3D with the albedo map.

As a future line of research, one can model the alignment between the clothing texture boundaries and the underlying geometry by learning a texture model that is coupled to shape.

\subsection{Pose-dependent clothing deformation}
In the clip from \texttt{03:18} in the supplemental video, we animate a test motion sequence of a clothed body. We fix the clothing structure variable $z$ and clothing type $c$, and generate new clothing offsets by only changing body pose $\theta$ (see main paper Sec.~\ref{sec:sample_experients}, ``Pose-dependent clothing deformation''). Then the clothed body is brought to animation with the corresponding pose.

We compare it with traditional rig-and-skinning methods with fixed
clothing offsets. An example of such a method is to dress a body with an instance of offset clothing layer using ClothCap \cite{pons2017clothcap}, and re-pose using SMPL blend skinning.

The result is shown in both the original motion and in the zero-pose space (\ie body is unposed to a ``T-pose''). In the zero-pose space, we exclude the pose blend shapes (body shape deformation that is caused by pose variation), to highlight the deformation of the clothes. As the rig-and-skinning method uses a single fixed offset clothing layer, it looks static in the zero-pose space. In contrast, the clothing deformation generated by CAPE is pose-dependent, temporal coherent, and more visually plausible.

\section{CAPE Dataset Details}\label{appendix:cape_dataset_details}
Elaborating on the main manuscript Sec.~\ref{sec:dataset},
our dataset consists of:
\begin{itemize}
\setlength\itemsep{0em}
 \item 40K registered 3D meshes of clothed human scans for each gender.
 \item 8 male and 3 female subjects.
 \item 4 different types of outfits, covering 8 common garment types: short T-shirts, long T-shirts, long jerseys, long-sleeve shirts, blazers, shorts, long pants, jeans.
 \item Large variations in pose.
 \item Precise, captured minimally clothed body shape.
\end{itemize}

Table \ref{table:public_dataset} shows a comparision with public 3D clothed human datasets. Our dataset is distinguished by accurate alignment, consistent mesh topology, ground truth body shape scans, and a large variation of poses. These features makes it not only suitable for studies on human body and clothing, but also for the evaluation of various Graph-CNNs. 
See Fig.~\ref{img:dataset_examples} and \texttt{01:56} in the supplemental video for examples of the dataset.
The dataset is available for research purposes at \websiteCAPE.

\begin{table*}[htb]
\small
\centering
\caption{Comparison with other datasets of clothed humans.}
\label{table:public_dataset}
\begin{tabular}{lcccccc}
\toprule
\multirow{2}{*}{\bf Dataset} & \multirow{2}{*}{\bf Captured} & \bf Body Shape & \multirow{2}{*}{\bf Registered} & \bf Large Pose & \bf Motion & \bf High Quality\\
 & & \bf Available &  & \bf Variation & \bf Sequences & \bf Geometry \\
\hline
Inria dataset~\cite{yang2016estimation} &Yes & Yes & No & No & Yes & No \\
BUFF \cite{Zhang_2017_CVPR} &Yes & Yes & No & No & Yes & Yes \\
Adobe dataset~\cite{vlasic2008articulated} &Yes & No & Yes$^*$ & No & Yes & No \\
RenderPeople  & Yes & No & No & Yes & No & Yes\\
3D People~\cite{pumarola20193dpeople}  &No & Yes & Yes$^*$ & Yes & Yes & Yes\\
\hline
\textbf{Ours} & Yes &  Yes & Yes & Yes & Yes & Yes\\
\bottomrule
\end{tabular}
\vspace{5pt}
\captionsetup{justification=raggedright,singlelinecheck=false}
\caption*{\qquad\qquad\quad$*$ Registered per-subject, \ie mesh topology is consistent only within the instances from the same subject.}
\vspace{-1.5em}
\end{table*}

\begin{figure*}[htb]
\vspace{10em}
\begin{center}
   \includegraphics[width=0.98\textwidth]{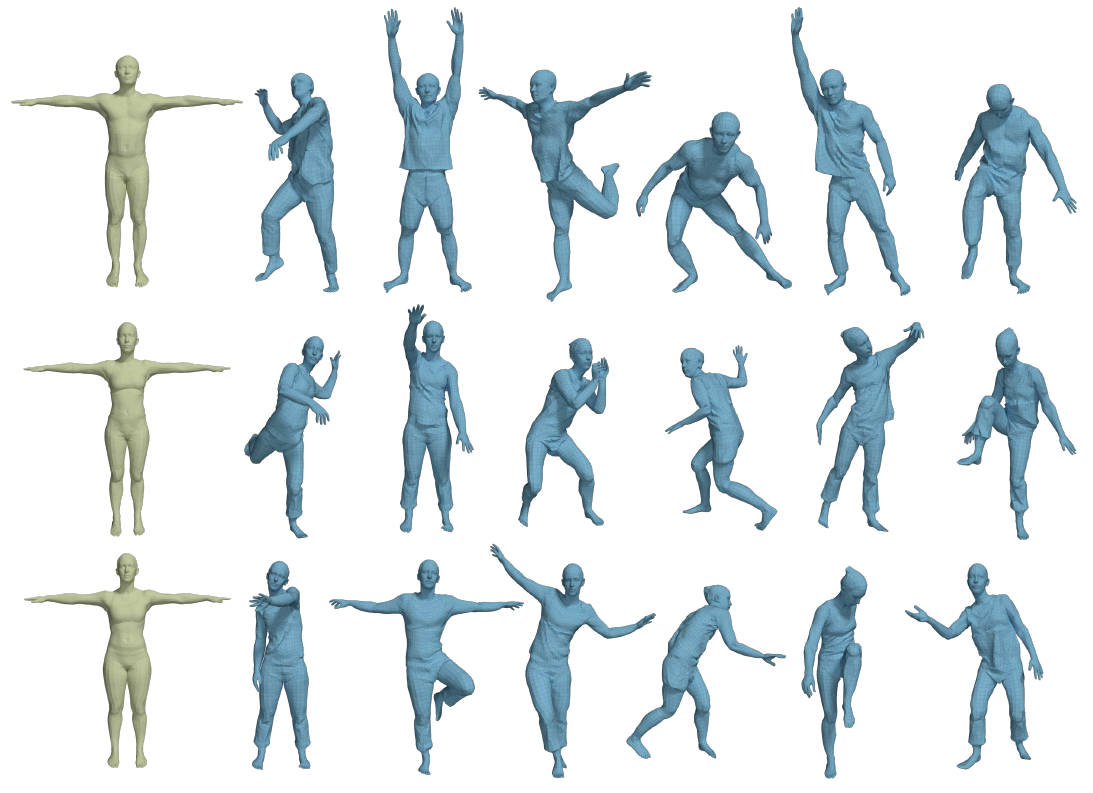}
\end{center}
\vspace{-1.2em}
\caption{Examples from the CAPE dataset: we provide accurate minimal-dressed body shape (green), clothed body scans with large pose and clothing wrinkle variations, all registered to the SMPL mesh topology (blue).}\label{img:dataset_examples}
\label{fig:dataset_examples}
\end{figure*}

\end{document}